%% file: main.tex
\documentclass{article}

\usepackage[preprint]{neurips_2026}

\usepackage[utf8]{inputenc}
\usepackage[T1]{fontenc}
\usepackage{hyperref}
\usepackage{url}
\usepackage{booktabs}
\usepackage{multirow}
\usepackage{amsfonts}
\usepackage{amsmath}
\usepackage{nicefrac}
\usepackage{microtype}
\usepackage{xcolor}
\usepackage{graphicx}

\graphicspath{{figures/}}

\title{Covariance-Aware Goodness for Scalable Forward-Forward Learning}

\author{
  Xiaoyi Jiang$^{1,2}$, Bashir M. Al-Hashimi$^{1}$, Kai Xu$^{1}$ \\
  $^{1}$King's College London, United Kingdom \\
  $^{2}$University of Nottingham \\
  \texttt{xiaoyi.jiang@kcl.ac.uk, bashir.al-hashimi@kcl.ac.uk, kai.xu@kcl.ac.uk}
}

\begin{document}

\maketitle

\begin{abstract}
\input{sections/abstract}
\end{abstract}

\input{sections/introduction}

\input{sections/related_work}
\input{sections/method}
\input{sections/experiments}
\input{sections/conclusion}

\begin{ack}
\input{sections/acknowledgments}
\end{ack}

\bibliography{refs}

\appendix
\input{appendix/appendix}

\end{document}

%% file: sections/abstract.tex
The Forward-Forward (FF) algorithm trains each layer using a local goodness objective, eliminating global gradient flow and full network activations storage. However, in convolutional settings, existing BP-free FF methods significantly under-perform backpropagation (BP) on complex benchmarks such as ImageNet-100 and Tiny-ImageNet. We identify this gap as a structural bottleneck in goodness extraction: standard sum-of-squares formulation collapses feature volumes into channel-wise activation energies, essentially first-order statistics that capture marginal feature strength (analogous to the diagonal of a covariance matrix). This process omits critical second-order dependencies that become increasingly discriminative at scale. To address this, we propose a framework centered on three key components. First, \emph{Bi-axis Covariance Goodness (BiCovG)} explicitly augments the standard goodness function with structured second-order information along two axes: cross-channel projections that model inter-feature covariance, and nested multi-scale aggregation that encodes spatial correlation statistics. This provides a tractable approximation to covariance-aware goodness without the prohibitive $O(C^2)$ complexity of explicit matrix estimation, yielding consistent per-layer gains of $+3$ to $+7\%$ and improving deep-layer utilization. Second, a lightweight Logistic Fusion module aggregates layer-wise predictions, amplifying the contribution of deeper representations. Third, the Feature Alignment Layer (FAL) introduces a zero-initialized correction at block boundaries to mitigate representation misalignment in deep locally trained networks. By introducing these three components, we effectively double the depth of viable Forward-Forward learning, extending robust layer utilization from shallow baselines to 16-layer architectures like VGG-16. The resulting BP-free model achieves $73.01\%$ on ImageNet-100 and $50.30\%$ on Tiny-ImageNet. As a practical extension, Hybrid Goodness Blocks (HGB) control the scope of gradient propagation via configurable block sizes, further narrowing the ImageNet-100 gap to $3.6\%$ ($83.98\%$) and matching BP on Tiny-ImageNet ($54.48\%$), while still reducing peak memory by approximately $50\%$ relative to BP.

%% file: sections/introduction.tex
\section{Introduction}
\label{sec:introduction}

Training deep networks via BP necessitates storing intermediate activations for the backward pass, causing peak memory to scale linearly with network depth. Local learning methods decouple this dependency by optimizing layers or blocks against local objectives, thereby bounding activation memory. Within this family, the Forward-Forward (FF) algorithm~\citep{hinton2022ff} offers a principled BP-free paradigm in which each layer is trained on a scalar \emph{goodness} extracted from its activations. While the original formulation~\citep{hinton2022ff} contrasts positive and negative samples, recent FF variants~\citep{cwc, deeperforward, asge} adopt local supervised objectives that map the goodness vector to class logits via cross-entropy; we follow this line. While subsequent work has extended FF to deeper convolutional architectures~\citep{cwc, deeperforward}, evaluations remain largely confined to CIFAR-scale datasets. As a result, the scalability of goodness-based local learning to more complex benchmarks such as ImageNet-100 remains insufficiently understood.

Existing FF goodness extraction has a structural limitation. The standard goodness for a feature tensor $F \in \mathbb{R}^{C \times H \times W}$ computes the squared energy per-channel: $g_c = \sum_{h,w} f_{c,h,w}^2$. Mathematically, this reduces the feature volume into a $C$-dimensional vector by aggregating each channel independently, retaining only marginal statistics. Statistically, this retains only the diagonal of the uncentered covariance matrix, representing individual channel variances, while discarding off-diagonal entries that encode cross-channel correlations. This reduction is lossy: discriminative information in convolutional representations often resides not in individual channel magnitudes, but in structured co-occurrence patterns across channels. Distinct classes may exhibit similar per-channel energies yet differ in their inter-channel correlations, which are eliminated by independent channel-wise aggregation. These limitations reflect a broader mismatch between the per-channel energy measure and the structure of convolutional features. Intermediate CNN layers learn channels corresponding to diverse visual patterns, and downstream classification depends on how these patterns interact. This is consistent with established results in computer vision: neural style transfer~\citep{gatys2016image} uses Gram matrices, a second-order cross-channel statistic, to capture perceptual structure, demonstrating the importance of off-diagonal covariance. Likewise, a standard classification head computes $W^\top f + b$, combining all channels into class-specific scores. In contrast, per-channel goodness discards the cross-channel dependencies that the classifier exploits.

Recent FF variants partially address channel structure, but not at the appropriate level. CwC~\citep{cwc} introduces class-wise channel grouping and enforces competition across groups through a softmax objective, injecting inter-class interaction into the loss. However, within each group, the goodness remains a sum of independent channel energies, which does not explicitly capture cross-channel correlations. Consequently, while inter-group competition is modeled, intra-group dependencies remain underutilized, potentially limiting the expressive capacity of the learned representations. Because this bottleneck is rooted in the constraints imposed on the feature extraction objective, it is not fully mitigated by longer training or increased depth. Its impact becomes more pronounced with task scale: the performance gap between BP-free FF and end-to-end BP increases from roughly $5\%$ on CIFAR-100 to approximately $28\%$ on ImageNet-100 ($224{\times}224$, 100 classes). This degradation is further compounded at depth by two factors: (i) a fixed random readout projection, whose capacity saturates as the number of classes and network depth grow, and (ii) representation misalignment across group boundaries in locally trained blocks, which hinders the formation of globally coherent features.

To address these issues, we propose a local learning framework that targets both representation and optimization bottlenecks. \textbf{Bi-axis Covariance Goodness (BiCovG)} generalizes goodness encoding along two orthogonal axes: cross-channel projections that capture off-diagonal covariance structure, and nested multi-scale spatial aggregation that encodes local correlation energy at varying granularities. Paired with a lightweight Logistic Fusion stage that aggregates predictions from all $L$ per-layer classifiers, BiCovG enhances deep-layer contributions and promotes more effective use of network depth. For deep grouped networks, the \textbf{Feature Alignment Layer (FAL)} applies a zero-initialized channel-wise correction at group boundaries, mitigating distributional shifts between locally trained blocks without disrupting greedy local optimization. Beyond the global BP-free setting, we further propose \textbf{Hybrid Goodness Blocks (HGB)}, a training scheme with configurable block size $m$ that provides a memory-efficient continuum between strict layer-wise learning and full backpropagation. Our BP-free model (BiCovG + FAL) achieves $73.01_{\pm0.49}\%$ accuracy on ImageNet-100 and $50.30_{\pm0.21}\%$ on Tiny-ImageNet, representing, to our knowledge, the strongest reported CNN-based global BP-free FF results on these benchmarks. As a practical extension, BiCovG-HGB ($4{\times}4$) attains $83.98_{\pm0.02}\%$ on ImageNet-100, narrowing the gap with end-to-end BP to $3.6\%$ while lowering peak GPU memory usage by $47.8\%$. \textbf{Our main contributions are as follows:} \textbf{1. Bi-axis Covariance Goodness (BiCovG)}: a bi-axis goodness encoding defined over the channel and spatial axes: learnable cross-channel projections model off-diagonal covariance across channels, while nested multi-scale spatial aggregation captures local correlation energy. Coupled with a lightweight Logistic Fusion stage, BiCovG yields consistent per-layer gains of $+3$ to $+7\%$, improving deep-layer utilization on ImageNet-scale benchmarks. \textbf{2. Feature Alignment Layer (FAL)}: a boundary-specific adaptation of residual adapter for gradient-isolated blocks, providing a zero-initialized channel-wise correction to mitigate cross-block representation misalignment in deep locally trained networks. \textbf{3. Hybrid Goodness Blocks (HGB)}: a training scheme with configurable block size that controls gradient propagation, bridging local BP-free learning and full backpropagation while retaining memory efficiency and reducing the performance gap.

%% file: sections/related_work.tex
\section{Related Work}
\label{sec:related_work}

\paragraph{BP-free and local learning.}
The landscape of training without end-to-end BP extends across several distinct families. Feedback alignment methods (e.g., FA~\citep{lillicrap2016fa,bartunov2018scalability}, DFA~\citep{nokland2016dfa,crafton2019dfa}) replace exact transpose weights with random or sparse feedback pathways to propagate error signals, while target-propagation (e.g., DTP~\citep{bengio2014targetprop,ernoult2022dtp}) optimizes layers against synthetic targets rather than global gradients. Biologically inspired local rules such as SoftHebb~\citep{moraitis2022softhebb,journe2023softhebb}, recursive local representation alignment (rLRA)~\citep{ororbia2023rlra}, and PEPITA~\citep{dellaferrera2023pepita}, which utilizes an error-modulated second forward pass, offer further alternatives to the backward pass. Despite these advancements, many necessitate specialized architectures or full forward-pass dependencies that differ from the scalable, goodness-based CNN training explored here. Local supervised learning frameworks like InfoPro~\citep{infopr} and AugLocal~\citep{auglocal} achieve ImageNet-scale viability via gradient-isolated blocks but still rely on internal BP and complex, multi-layer auxiliary classifiers.

\paragraph{Forward-Forward methods.}
The Forward-Forward algorithm~\citep{hinton2022ff} introduced per-layer goodness maximization as a BP-free training paradigm. Subsequent FF variants have improved either sample construction or layer-wise supervision: CaFo~\citep{zhao2023cafo} cascades local predictors on a forward-only backbone, while SymBa~\citep{lee2023symba} symmetrizes the contrastive objective. Following recent FF works, each block is trained using a local cross-entropy loss against the ground-truth label. CwC~\citep{cwc} adopts class-wise channel grouping with competitive goodness, and DeeperForward~\citep{deeperforward} stabilizes deeper FF via normalization and parallel layer-wise optimization; however, both encounter channel explosion as class count grows. ASGE~\citep{asge} decouples goodness dimensionality from class count via adaptive spatial encoding, yet per-layer accuracy still degrades with depth, restricting it to shallow architectures such as VGG8. SCFF~\citep{scff} improves negative sample generation for self-contrastive FF training, and CFF+M~\citep{cff} adapts FF-style training to Vision Transformers. Despite this progress, most FF results remain concentrated on CIFAR-scale tasks or different architecture families, and systematic BP-free FF evaluation on ImageNet-100 and Tiny-ImageNet remains limited.

\paragraph{Goodness encoding and feature representation.}
The goodness function in existing FF paradigms relies on variants of per-channel spatial energy, calculated either globally~\citep{hinton2022ff}, within class-aligned groups~\citep{cwc}, or over adaptive spatial partitions~\citep{asge}. Mathematically, these formulations reduce the feature volume to marginal second moments, effectively isolating diagonal covariance information while omitting inter-channel correlations. While cross-channel dependencies are central to modern supervised architectures, facilitated by channel attention~\citep{hu2018senet} and mixing layers, they remain underexplored within the local goodness objective. Furthermore, while multi-scale spatial pooling~\citep{he2015spp} is standard for handling scale variance in BP-based vision, FF goodness has typically been computed at a single spatial granularity. BiCovG bridges this gap by unifying these axes into a single local objective: learnable projections probe off-diagonal covariance structure, while nested multi-scale aggregation captures spatial co-activation statistics. To our knowledge, BiCovG is the first goodness encoding to move beyond diagonal covariance structure on both channel and spatial axes simultaneously.

%% file: sections/method.tex
\section{Method}
\label{sec:method}

We propose a BP-free local learning framework for CNNs that replaces global gradients with five synergistic components: (1) \textbf{goodness-based local supervision}, enabling independent layer optimization without inter-layer gradient flow; (2) \textbf{Bi-axis Covariance Goodness (BiCovG)}, which moves beyond diagonal energy to capture cross-channel correlations and multi-scale dependencies; (3) a \textbf{Feature Alignment Layer (FAL)} that mitigates representation misalignment across decoupled block boundaries; (4) \textbf{Logistic Fusion}, a lightweight strategy for aggregating multi-level predictions into a global inference; and (5) \textbf{Hybrid Goodness Blocks (HGB)}, which generalize the framework to configurable gradient scopes, balancing memory efficiency with optimization depth.

\subsection{Goodness-Based Local Supervision}
\label{sec:method:local}

The Forward-Forward (FF) algorithm~\citep{hinton2022ff} optimizes each layer independently to maximize a scalar \emph{goodness} metric for positive data while minimizing it for negative samples, eliminating the need for backpropagation across layers. We extend this local supervision principle by formalizing goodness extraction as a per-layer supervised classification objective. For each convolutional block $l$, we compute a multi-dimensional goodness vector $\mathbf{g}_l^{\text{BiCovG}} \in \mathbb{R}^D$ from post-ReLU activations (see Section~\ref{sec:method:bicovg}). This representation is mapped to class logits via a trainable linear readout:
\begin{equation}
    \hat{\mathbf{y}}_l = W_l\,\mathbf{g}_l^{\text{BiCovG}} + \mathbf{b}_l
\end{equation}
where $W_l \in \mathbb{R}^{K \times D}$ and $\mathbf{b}_l \in \mathbb{R}^K$ are task-specific learnable parameters, with $W_{ij},\, b_j \sim \mathrm{Uniform}(-1/\sqrt{D},\, 1/\sqrt{D})$, where $D$ is the goodness dimension. In contrast to prior FF methods~\citep{hinton2022ff}, which rely on \emph{fixed} random projections, this learnable $W_l$ removes a key structural bottleneck by allowing each layer to adaptively form the most discriminative linear combinations of goodness features, improving scalability with increasing depth and class counts. Each block is trained using a local cross-entropy loss $\mathcal{L}_l = \operatorname{CE}(\hat{\mathbf{y}}_l, y)$ against the ground-truth label. The parameters $\theta_l$ of block $l$ receive gradients exclusively from $\mathcal{L}_l$ with no gradient propagation across block boundaries. The total loss $\mathcal{L}_{\text{total}} = \sum_{l=0}^{L-1} \mathcal{L}_l$ is used for monitoring purposes only.

\subsection{Bi-Axis Covariance Goodness (BiCovG)}
\label{sec:method:bicovg}

\begin{figure*}[t]
    \centering
    \includegraphics[width=\textwidth]{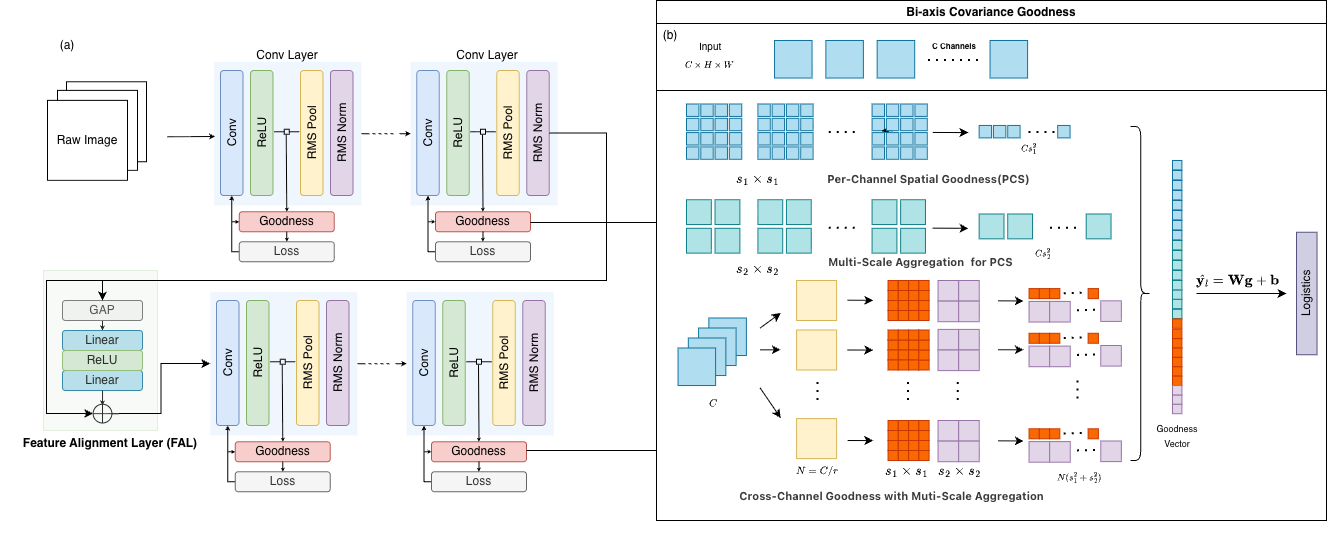}
    \caption{\textbf{(a) Network architecture.} Each block produces post-ReLU activations for BiCovG; outputs are detached between blocks. Feature Alignment Layers (FAL) at group boundaries correct representation misalignment. \textbf{(b) BiCovG.} At two spatial scales ($s_1 < s_2$), per-channel spatial (PCS) goodness and cross-channel (CC) goodness capture complementary representations. The resulting features are concatenated into $\mathbf{g}_l^{\text{BiCovG}}$ and mapped to class logits via a learnable readout $W_l$.}
    \label{fig:overview}
\end{figure*}

\paragraph{Motivation.}
Recent FF methods compute goodness as per-channel or group-channel spatial energy~\citep{cwc, deeperforward}, capturing only channel-wise marginal activation statistics. Concretely, per-channel energy $g_{l,c} = \mathbb{E}_{h,w}[f_{l,c,h,w}^2]$ measures activation magnitude independently for each channel, without modeling cross-channel correlations. As a result, jointly informative co-activation patterns between channels are not explicitly captured. 
On the spatial axis, this formulation collapses each $H \times W$ feature map into a single scalar per channel via global pooling. This operation discards the spatial distribution and scale-variant characteristics of the activations, effectively treating the feature map as a spatially stationary signal. However, the spatial co-activation patterns, i.e., the second-order dependencies between neighboring positions and their evolution across scales, carry critical discriminative information that is rendered invisible by global aggregation.

Bi-axis Covariance Goodness (BiCovG, Figure~\ref{fig:overview}) addresses these limitations by extending the goodness representation along two orthogonal axes: (1) \textbf{Cross-channel projections} that probe off-diagonal covariance structure to capture inter-feature interactions, and (2) \textbf{Nested multi-scale spatial aggregation}, which hierarchically computes energy across varying granularities to expose spatial second-order structure.

\paragraph{Per-channel spatial goodness.}
For a discrete scale parameter $s$, we uniformly partition the spatial domain $H \times W$ into a grid of $s^2$ non-overlapping regions $\mathcal{R}_{ij}^{s}$, where $i,j \in \{1,\ldots,s\}$. Within each region, the per-channel spatial goodness energy is:
\begin{equation}
    g_{b,c,i,j}^{\text{pcs}}(s) = \frac{1}{|\mathcal{R}_{ij}^{s}|} \sum_{(h,w)\in\mathcal{R}_{ij}^{s}} f_{b,c,h,w}^2
\end{equation}
where $f_{b,c,h,w}$ is the post-ReLU activation at batch index $b$, channel $c$, and spatial coordinate $(h,w)$. Concatenating these local energies across all channels and regions yields $\mathbf{g}^{\text{pcs}}(s) \in \mathbb{R}^{B \times Cs^2}$.

\paragraph{Cross-channel goodness.}
Let $W^{\text{cc}} \in \mathbb{R}^{N \times C}$ be a learnable projection matrix where $N = C/r$ (with reduction ratio $r = 8$). For each spatial region $\mathcal{R}_{ij}^s$:
\begin{equation}
    g_{b,k,i,j}^{\text{cc}}(s) = \frac{1}{|\mathcal{R}_{ij}^{s}|}\sum_{(h,w)\in\mathcal{R}_{ij}^{s}} \left(\sum_{c=1}^{C} W^{\text{cc}}_{kc} \cdot f_{b,c,h,w}\right)^{2}
\end{equation}
Each row $\mathbf{w}_k^{\text{cc}}$ linearly mixes all $C$ channels into a scalar before squaring, yielding $N$ complementary energy measurements that capture cross-channel interactions. The resulting vector is $\mathbf{g}^{\text{cc}}(s) \in \mathbb{R}^{B \times Ns^2}$. This branch is structurally akin to a $1{\times}1$ convolution followed by a region-averaged squared-energy. Ablation on $r \in \{4, 8, 16\}$ shows negligible difference between $r=4$ and $r=8$ ($<0.5\%$ on both Tiny-IN and IN-100), while $r=16$ trails by $0.9\%$ and $1.1\%$, respectively. We therefore adopt $r=8$ as a practical balance between accuracy, parameter cost, and computation.

Unlike CwC~\citep{cwc}, our method avoids channel-capacity collapse caused by rigid class-based channel partitioning. Instead of assigning channels to classes, we learn linear projections across channels, effectively sampling the channel covariance structure $\Sigma_\Omega \in \mathbb{R}^{C \times C}$ with $N = C/r$ directions $\{\mathbf{w}_h\}$. Given the redundancy of CNN feature channels, we assume that the discriminative structure of $\Sigma_\Omega$ concentrates in a low-dimensional subspace; therefore, $C/r$ linearly independent directions are sufficient to capture the relevant inter-channel correlation information.

\paragraph{Multi-scale aggregation.}
We apply two spatial scales $s_1 < s_2$ and concatenate the components:
\begin{equation}
    \mathbf{g}_l^{\text{BiCovG}} = \left[\,\mathbf{g}^{\text{pcs}}(s_1)\;\big|\;\mathbf{g}^{\text{cc}}(s_1)\;\big|\;\mathbf{g}^{\text{pcs}}(s_2)\;\big|\;\mathbf{g}^{\text{cc}}(s_2)\,\right]
\end{equation}
Scale $s_1$ aggregates correlation energy over broader regions, while scale $s_2$ resolves it at a finer spatial granularity. Together they yield spatial second-order statistics at two granularities, complementing cross-channel covariance probes. Scale pairs are selected per layer group to keep goodness dimensionality consistent, despite varying channel counts or spatial resolutions (see Section~\ref{sec:method:arch}).

\subsection{Network Architecture and Per-Layer BiCovG Integration}
\label{sec:method:arch}

The framework utilizes a 16-layer convolutional backbone following the VGG-16~\citep{simonyan2015vgg} architecture. Each block is optimized for local, gradient-isolated training and follows a standardized sequence:
\begin{equation}
    \mathbf{f}_l = \text{ReLU}\!\left(\text{Conv}_{3\times3}(\mathbf{h}_{l-1})\right), \qquad
    \mathbf{h}_l = \text{RMSNorm}\!\left(\text{Dropout}\!\left(\text{Pool}(\mathbf{f}_l)\right)\right)
\end{equation}
where $\mathbf{h}_{l-1}$ is the detached output of the preceding block. The post-ReLU activation $\mathbf{f}_l$ serves as the primary input for BiCovG computation, while $\mathbf{h}_l$ is the normalized output forwarded to the subsequent block. We employ \textbf{RMSNorm}~\citep{zhang2019rmsnorm}, which normalizes activation magnitudes without learnable affine parameters: $\text{RMSNorm}(\mathbf{x}) = \mathbf{x} / (\sqrt{\operatorname{mean}(\mathbf{x}^2)} + \epsilon)$, where the mean is taken over all channels and spatial dimensions. For spatial downsampling at four designated layers, we utilize \textbf{RMSPool} rather than standard max or average pooling: $\text{RMSPool}(\mathcal{R}) = \sqrt{\frac{1}{|\mathcal{R}|}\sum_{(h,w)\in\mathcal{R}}\bigl(\mathcal{R}^{(h,w)}\bigr)^{2}}$. This preserves the energy-based relationship that BiCovG relies on.

\begin{table}[ht]
\centering
\small
\caption{Per-layer BiCovG integration. Scale pairs $(s_1, s_2)$ are selected per channel tier to maintain uniform goodness dimensionality. Spatial sizes shown as ranges for Tiny-ImageNet ($64^2$) and ImageNet-100 ($224^2$ input, downsampled to $112^2$ at layer 0).}
\label{tab:bicovg_integration}
\begin{tabular}{ccccc}
\toprule
Layer group & $C$ & Spatial (Tiny-IN / IN-100) & BiCovG scales $(s_1, s_2)$ & BiCovG dim \\
\midrule
0--1  (shallow) & 128     & $64{\to}32$ / $112{\to}56$ & $(2,\,4)$ & 2880 \\
2--7  (mid)     & 256/512 & $16$--$32$ / $28$--$56$    & $(1,\,2)$ & 1440/2880 \\
8--15 (deep)    & 512     & $4$--$8$ / $7$--$14$       & $(1,\,2)$ & 2880 \\
\bottomrule
\end{tabular}
\end{table}

Table~\ref{tab:bicovg_integration} summarizes the per-layer BiCovG configuration; scale pairs are chosen to maintain uniform goodness dimensionality (${\approx}2880$) across all layer groups. The complete 16-layer architecture is provided in Appendix~\ref{app:arch}.

\subsection{Feature Alignment Layer (FAL)}
\label{sec:method:fal}

Decoupling the network into $G$ gradient-isolated groups can introduce representation misalignment at group boundaries: the output of one independently optimized block may be poorly conditioned as the input to the next, forcing downstream groups to expend capacity on adaptation. To correct this, we introduce FAL, a boundary-specific adaptation of residual adapters~\citep{rebuffi2017adapters, houlsby2019adapter} for the layer-wise FF setting. FAL applies a channel-wise additive correction:
\begin{equation}
    \mathbf{h}_l' = \mathbf{h}_l + \boldsymbol{\Delta} \otimes \mathbf{1}_{H \times W},
    \qquad
    \boldsymbol{\Delta} = W_2\,\text{ReLU}(W_1\,\text{GAP}(\mathbf{h}_l)),
\end{equation}
where $\boldsymbol{\Delta} \in \mathbb{R}^C$ is shared across spatial positions and $W_1, W_2$ form a fixed hidden dimension $d = 512$. Zero-initializing $W_2$ ensures FAL starts as the identity and learns its correction purely from data; its parameters are optimized by the local loss of the first layer of the subsequent group, preserving the greedy local learning property.

\subsection{Training: Greedy Local Learning and Logistic Fusion}
\label{sec:method:training}
Our training has two stages: \emph{Greedy local backbone training} and \emph{Logistic Fusion}. In Stage 1, each block's output is detached from the computational graph before forwarding, ensuring $\theta_l$ is optimized exclusively by the local loss
$\mathcal{L}_l$. This realizes a gradient-isolated Decoupled Greedy Learning scheme~\citep{pmlr-v119-belilovsky20a} that avoids storage of intermediate activations. Once the backbone is frozen, we perform \emph{Logistic Fusion} on the same training split. While greedy local training distributes complementary information across layers, relying on 
a single layer's prediction discards potentially useful hierarchical features. We aggregate layer-wise logits $\{\hat{\mathbf{y}}_l\}$ via a lightweight head:
\begin{equation}
    \hat{\mathbf{y}}_{\text{fused}} = \sum_{l=0}^{L-1} w_l\,\hat{\mathbf{y}}_l,
    \qquad \mathbf{w} = \operatorname{softmax}(\boldsymbol{\alpha})
\end{equation}
where $\boldsymbol{\alpha} \in \mathbb{R}^L$ is trained with cross-entropy
(Adam~\citep{kingma2015adam}, lr $= 0.01$, 500 epochs). The fusion introduces only
$L = 16$ scalar parameters and trains in under one minute on a single GPU.
We report two inference modes: \textbf{Best Pred}, a zero-parameter baseline using the logits of the single most accurate block, and \textbf{Fusion Pred}, which applies the learned fusion to explore cross-layer complementarity.

\subsection{Hybrid Goodness Block (HGB)}
\label{sec:method:hgb}

Strict layer-wise training ($m{=}1$) represents the purest FF setting, preserving complete local independence that motivates FF. Inspired by gradient-isolated local supervised learning~\citep{infopr,auglocal}, where small blocks of layers are trained jointly via internal BP under an auxiliary objective, we transplant the concept of inter-block BP into the goodness-based regime. HGB partitions the network into groups of $m$ consecutive layers, where each group is optimized jointly via standard BP against a single goodness objective at the block's exit. Crucially, activations are detached between groups, and the FAL preserves this cross-block independence. Within each block, BiCovG, FAL, and the readout $W_l$ are instantiated only at the boundary layer, while intermediate layers utilize standard convolutional operations. This yields $L/m$ logits that feed into Logistic Fusion. This defines a tunable continuum of gradient horizon: $m=1\ (\text{strictly layer-wise, BP-free}) \ \longrightarrow\ m=2,4\ (\text{HGB}) \ \longrightarrow\ m=L\ (\text{full BP})$, along which accuracy can be recovered by selectively widening the gradient horizon without paying the full activation-memory cost of end-to-end BP.

%% file: sections/experiments.tex
\section{Experiments}
\label{sec:experiments}

\subsection{Experimental Setup}
\label{sec:exp:setup}

We evaluate on three benchmarks of increasing scale: \textbf{CIFAR-100} ($32{\times}32$, 100 classes), \textbf{Tiny-ImageNet}~\citep{tinyimagenet} (\textbf{Tiny-IN}; $64{\times}64$, 200 classes), and \textbf{ImageNet-100}~\citep{russakovsky2015imagenet} (\textbf{IN-100}; $224{\times}224$, 100 classes). BP-free FF results on these benchmarks remain limited in the literature. Tiny-IN and IN-100 use a VGG-16 network (Section~\ref{sec:method:arch}), while CIFAR-100 uses a shallower 8-block variant. ImageNet-scale datasets are trained via SGD for 90 epochs (cosine decay, $0.05{\to}5{\times}10^{-4}$), while CIFAR-100 uses AdamW for 400 epochs. Full details are in Appendix~\ref{app:training}. Tiny-IN and IN-100 results represent the mean$\,\pm\,$std across three seeds. We report accuracy for both \textbf{Best Pred} (best single-layer) and \textbf{Logistic Fusion} (Section~\ref{sec:method:training}) inferences.
\subsection{Results and Ablation}
\label{sec:exp:ablation_study}

\begin{table}[ht]
\centering
\small
\caption{Ablation of BiCovG along both dimensions (Logistic Fusion, single seed; FAL excluded).}
\label{tab:bicovg_ablation}
\begin{tabular}{lcc}
\toprule
Configuration & Tiny-IN & IN-100 \\
\midrule
Per-channel goodness         & 43.25 & 64.87 \\
\quad$+$ Cross-Channel (CC)  & 46.12 & 67.16 \\
\quad$+$ Multi-Scale (MS)    & 47.91 & 68.55 \\
BiCovG                          & 50.38 & 71.60 \\
\bottomrule
\end{tabular}
\end{table}

Table~\ref{tab:bicovg_ablation} decomposes BiCovG along its two axes (FAL excluded). Relative to per-channel goodness, cross-channel correlation (CC) probes and multi-scale spatial partitioning (MS) provide complementary gains: on Tiny-IN/IN-100, CC yields $+2.87$/$+2.29\%$, MS yields $+4.66$/$+3.68\%$, and their combination reaches $+7.13\%$/$+6.73\%$, closely matching the sum of individual contributions. Adding FAL further improves IN-100 by $+1.4\%$, with a marginal change on Tiny-IN ($-0.08\%$, within noise) while still producing measurable per-layer improvement (Section~\ref{sec:exp:ablation}).

Table~\ref{tab:main_results} compares our method with prior FF and BP-free baselines, and includes key ablation variants. All Tiny-IN and IN-100 results for our models use Logistic Fusion.
\begin{table}[ht]
\centering
\small
\setlength{\tabcolsep}{3pt}
\caption{Comparison with BP-free and FF-based methods. ``--'' denotes missing results in the original work. Tiny-IN and IN-100 results for ASGE (repr.) and our models are reproduced or measured under our protocol. For our models, Tiny-IN and IN-100 entries are reported as Best Pred / Logistic Fusion. $\dagger$: CFF+M uses ViT and is not directly comparable to CNN-based methods. Bold indicates best within each group per column.}
\label{tab:main_results}
\resizebox{\textwidth}{!}{%
\begin{tabular}{llccc}
\toprule
Method & Model & CIFAR-100 & Tiny-ImageNet & ImageNet-100 \\
\midrule
\multicolumn{5}{l}{\textit{Non-FF BP-free}} \\
FA            & MLP8           & -- & -- & -- \\
DTP                 & VGG6           & -- & -- & -- \\
DFA                     & CNN5           & $41.00{\pm}0.30$ & -- & -- \\
SoftHebb        & SoftHebb       & $\mathbf{56.00}$ & -- & -- \\
rLRA                                    & ResNet18       & -- & -- & -- \\
PEPITA                        & CNN2           & $27.56{\pm}0.60$ & -- & -- \\
\midrule
\multicolumn{5}{l}{\textit{FF-based local-learning}} \\
FF   & MLP4           & -- & -- & -- \\
CaFo  & CNN3           & $40.76$ & -- & -- \\
SymBa  & MLP3           & $29.28$ & -- & -- \\
CwC  & CNN4           & $51.23$ & -- & -- \\
DeeperForward  & ResNet18       & $53.09$ & -- & -- \\
ASGE   & VGG8           & $65.34$ & -- & -- \\
ASGE (repr.)  & VGG-16          & -- & $33.17$ & $\mathbf{59.38}$ \\
SCFF  & AlexNet-style CNN5 & -- & $35.67$ & -- \\
NoProp  & NoProp-DT      & $46.06$ & $25.65$ & -- \\
CFF+M$^\dagger$                                     & ViT            & $\mathbf{80.39}$ & $\mathbf{73.23}$ & -- \\

\textbf{BiCovG }(no FAL) & VGG8 & $68.21$ & -- & --            \\
\textbf{BiCovG }+ FAL    & VGG8 & $\mathbf{70.80}$ & -- & -- \\
\textbf{BiCovG} (no FAL) & VGG-16 & -- & $48.20{\pm}0.41$ / $\mathbf{50.38{\pm}0.42}$ & $70.83{\pm}0.13$ / $71.60{\pm}0.29$          \\
\textbf{BiCovG }+ FAL & VGG-16 & -- &$48.41{\pm}0.21$ / $50.30{\pm}0.21$ & $72.88{\pm}0.80$ / $\mathbf{73.01{\pm}0.49}$ \\

\midrule
\multicolumn{5}{l}{\textit{Ours -- HGB (block-internal BP)}} \\
\textbf{BiCovG-HGB} ($2{\times}8$, $m=2$)       & VGG-16 & -- &$48.89{\pm}0.40$ / $50.20{\pm}0.19$          & $77.95{\pm}0.32$ / $78.22{\pm}0.14$          \\
\textbf{BiCovG-HGB }($4{\times}4$, $m=4$)       & VGG-16 & -- & $52.42{\pm}0.28$ / $\mathbf{54.48{\pm}0.08}$ & $83.79{\pm}0.02$ / $\mathbf{83.98{\pm}0.02}$ \\
\midrule
\multicolumn{5}{l}{\textit{BP End-to-End}} \\
BP (repr.)                     & VGG8 & $\mathbf{70.22}$  & --         & --       \\
BP (repr.)                     & VGG-16 & --  & $\mathbf{54.17{\pm}0.37}$ & $\mathbf{87.61{\pm}0.17}$       \\
\bottomrule
\end{tabular}
}
\end{table}
Our BP-free model (BiCovG + FAL) achieves $73.01_{\pm0.49}\%$ on IN-100 and $50.30_{\pm0.21}\%$ on Tiny-IN, substantially outperforming prior FF methods (e.g., SCFF at $35.67\%$ on Tiny-IN). On CIFAR-100, it reaches $70.80\%$, slightly exceeding the BP baseline ($70.22\%$). Logistic Fusion consistently outperforms Best Pred, indicating complementary information across layers. With HGB, increasing block size $m$ progressively narrows the gap to full BP. At $m{=}4$, performance matches BP on Tiny-IN ($54.48$ vs.\ $54.17$) and approaches it on IN-100 ($83.98\%$, within $3.6\%$ of BP). At $m{=}2$, the shorter BP chain is more sensitive to cross-block boundary misalignment, yielding negligible gains on Tiny-IN but still $+5.21\%$ on IN-100. All HGB settings retain substantial memory savings (Table~\ref{tab:efficiency}).

\subsection{Mechanism and Efficiency Analysis}
\label{sec:exp:ablation}

We analyze per-layer behavior using four metrics derived from the accuracy curve $\{\text{acc}[l]\}_{l=0}^{L-1}$. The \textbf{Decline Area} measures post-peak degradation as the cumulative drop from the peak layer, capturing both the rate and the span of degradation.
Let $l^*$ denote the peak layer; we then define $\text{DA} = \sum_{l > l^*} \max\!\bigl(0,\;\text{acc}[l^*] - \text{acc}[l]\bigr)$. \textbf{Tail Retention} is given by $\text{TR} = \operatorname{mean}(\text{acc}[L{-}4\,:\,L{-}1])\,/\,\text{acc}[l^*]$, which quantifies the fraction of peak accuracy preserved in the final four layers. Values near 1 indicate no deep-layer degradation. \textbf{Shallow Gain (SG)} and \textbf{Deep Gain} track the mean per-layer improvements (layers 0--7 vs.\ layers 8--15) between configurations (e.g., A vs.\ B):
\begin{equation}
    \text{SG} = \frac{1}{8}\sum_{l=0}^{7}\!\bigl(\text{acc}_B[l]-\text{acc}_A[l]\bigr), \qquad
    \text{DG} = \frac{1}{8}\sum_{l=8}^{15}\!\bigl(\text{acc}_B[l]-\text{acc}_A[l]\bigr)
\end{equation}
where the split at $l=8$ divides the 16 layers into shallow (0--7) and deep (8--15) halves (Table~\ref{tab:bicovg_integration}). Using learned Logistic Fusion weights $\{w_l\}$, we compute a participation ratio to estimate how many layers contribute meaningfully to the fused prediction: $N_\text{eff} = {\bigl(\sum_l w_l\bigr)^2}/{\sum_l w_l^2}$ with $N_\text{eff} = 1$ when all weight is concentrated on a single layer and $N_\text{eff} = L$ when weights are uniform.

\begin{table}[ht]
\centering
\small
\caption{Diagnostic metric changes for BiCovG (per-channel goodness $\to$ BiCovG, holding architecture and training fixed) and FAL (added on top of BiCovG). Each column pair (Tiny-IN / IN-100) is averaged over controlled ablation pairs. Arrows indicate the direction of improvement.}
\label{tab:ablation_metrics}
\begin{tabular}{l ccc ccc}
\toprule
\multirow{2}{*}{Metric} &
  \multicolumn{3}{c}{per-ch.\,$\to$\,BiCovG $(\uparrow)$} &
  \multicolumn{3}{c}{$+$FAL $(\uparrow)$} \\
\cmidrule(lr){2-4}\cmidrule(lr){5-7}
 & Tiny-IN & IN-100 & Avg & Tiny-IN & IN-100 & Avg \\
\midrule
SG (\%) $\uparrow$ & $+2.2$ & $+2.7$ & $+2.45$ & $+0.74$ & $+1.39$ & $+1.06$ \\
DG (\%) $\uparrow$ & $+4.4$ & $+3.4$ & $+3.90$ & $+1.49$ & $+1.84$ & $+1.67$ \\
DA $\Delta$ $\downarrow$ & $-8.7$  & $-4.5$  & $-6.6$  & $+3.9$  & $+2.6$  & $+3.2$ \\
TR $\Delta$ $\uparrow$ & $+0.050$ & $+0.019$ & $+0.035$ & $+0.005$ & $-0.008$ & $-0.002$ \\
$N_\text{eff}$ $\Delta$ $\uparrow$ & $+0.30$ & $+1.16$ & $+0.73$ & $-0.02$ & $+0.17$ & $+0.08$ \\
\bottomrule
\end{tabular}
\end{table}

\begin{figure}[ht]
\centering
\includegraphics[width=\linewidth]{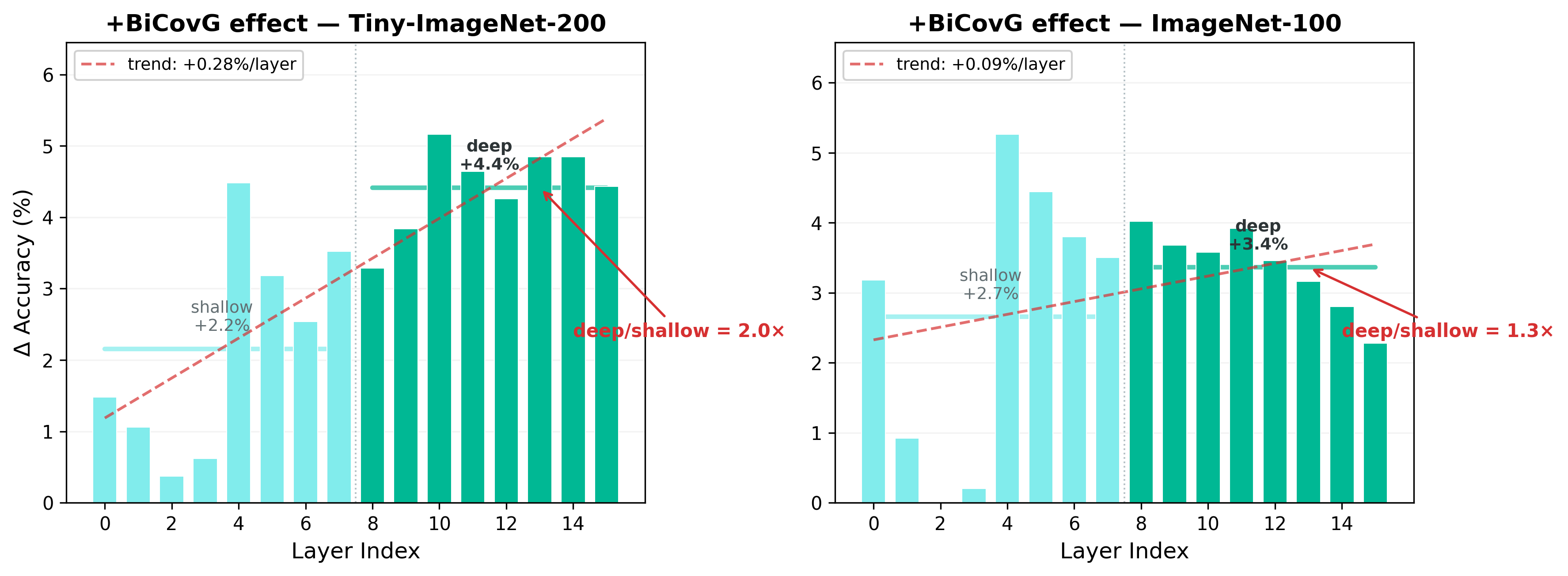}
\caption{Per-layer accuracy gain from replacing per-channel goodness with BiCovG on Tiny-IN (left) and IN-100 (right). The dashed line shows a consistent deep $>$ shallow trend; gains are most pronounced immediately after spatial downsampling (RMSPool), where BiCovG's multi-scale partitioning captures the restructured feature layout.}
\label{fig:bicovg_delta}
\end{figure}

BiCovG yields concurrent gains in both Shallow Gain ($+2.45\%$) and Deep Gain ($+3.90\%$), indicating that the cross-channel and multi-scale extensions bring additional discriminative information across the network. The most consequential effect is on post-peak degradation: DA is reduced by $6.6$ (with TR improving by $+0.035$ and $N_\text{eff}$ rising by $+0.73$), indicating that BiCovG keeps the goodness signal discriminative further into the network, effectively extending the depth at which local supervision remains productive. FAL operates on a different axis: it does not reduce DA, yet yields consistent per-layer gains (SG $+1.06\%$, DG $+1.67\%$), suggesting it redistributes the representational budget at group boundaries without altering the underlying goodness feature space.

\begin{table}[ht]
\centering
\small
\caption{Peak GPU memory on ImageNet-100 (VGG-16). Left: single GPU, batch size 128. Right: DDP 2-GPU, batch size 64/GPU (effective 128). Rows show representative configurations.}
\label{tab:efficiency}
\begin{tabular}{lrcc rcc}
\toprule
 & \multicolumn{3}{c}{Single GPU (A100)} & \multicolumn{3}{c}{DDP 2-GPU (A100)} \\
\cmidrule(lr){2-4}\cmidrule(lr){5-7}
Model & Params & Peak Mem & vs BP & Params & Peak/GPU & vs BP \\
\midrule
BP (VGG-16)                   & 134.7M & 16.75 GB & —       & 134.7M & 10.11 GB & — \\
BiCovG + FAL                     &  34.9M & 10.18 GB & $-39.2\%$ &  34.9M &  5.41 GB & $-46.5\%$ \\
BiCovG-HGB ($2{\times}8$)        &  31.2M &  8.94 GB & $-46.6\%$ &  31.2M &  4.87 GB & $-51.8\%$ \\
BiCovG-HGB ($4{\times}4$)        &  29.5M &  8.74 GB & $-47.8\%$ &  29.5M &  4.74 GB & $-53.1\%$ \\
\bottomrule
\end{tabular}
\end{table}

BP-free FF training improves efficiency on two axes. First, replacing VGG-16's FC head with per-layer linear readouts $W_l$ compresses total parameters from $134.7\text{M}$ to $34.9\text{M}$. Second, peak memory drops by $39$--$53\%$ versus standard BP. While the smaller parameter footprint accounts for ${\sim}1.2$ GB of this reduction, the dominant ${\sim}5.4$ GB saving stems from eliminating inter-block activation storage under gradient isolation. 
Within HGB, peak memory varies with $m$ in a U-shaped manner. This is driven by a trade-off between the number of per-step goodness backward computations (scaling as $L/m$) and the depth of internal activation within a block (scaling as $m$). At $m{=}1$, the simultaneous resident memory of $L$ layers' goodness graphs creates the peak; at $m{=}L$, full-network activations dominate. Intermediate settings balance these effects, with $m{=}4$ performing best among tested configurations. HGB retains these efficiency benefits while improving accuracy. 
Training throughput on a single RTX 5090 is comparable (BP: $559$ img/s, BiCovG-HGB ($4{\times}4$): $551$ img/s, and BiCovG + FAL: $550$ img/s). The block-independence of BP-free training further enables parallel execution, with up to $5.22\times$ theoretical speedup (one block per device), while HGB ($4{\times}4$) achieves $1.88\times$ with four groups. Interleaved sequential execution (Appendix~\ref{app:interleaved}) complements this by reducing peak activation memory from $O(L)$ to $O(1)$ without altering the training objective.

%% file: sections/conclusion.tex
\section{Conclusion}
\label{sec:conclusion}

This work identifies goodness extraction as a key bottleneck in scaling BP-free Forward-Forward learning. \textbf{Bi-axis Covariance Goodness (BiCovG)} augments goodness encoding along two orthogonal axes: cross-channel projections that probe off-diagonal covariance and nested multi-scale spatial aggregation. Per-layer analysis confirms that BiCovG mitigates deep-layer saturation, particularly in late-stage blocks where traditional per-channel goodness plateaus. When combined with Logistic Fusion, BiCovG increases the number of layers that contribute meaningfully to the final prediction, effectively leveraging the model's full hierarchical depth. Furthermore, the \textbf{Feature Alignment Layer (FAL)} introduces a lightweight boundary correction that improves representation consistency between gradient-isolated groups. Together, these components achieve $73.01\%$ on ImageNet-100 and $50.30\%$ on Tiny-ImageNet, outperforming prior BP-free baselines by a significant margin. As a practical bridge, \textbf{Hybrid Goodness Blocks (HGB)} allow for a tunable gradient scope, narrowing the gap to end-to-end BP while maintaining an approximately $50\%$ reduction in peak GPU memory.

Two limitations provide avenues for future research. First, evaluating BiCovG on ImageNet-1K will further test our cross-channel and multi-scale hypotheses at higher resolution and class granularity. Second, FAL's impact varies by dataset; characterizing this variance against spatial resolution, depth, and class granularity, and extending HGB's gradient-scope concept to mixed-precision and neuromorphic hardware, are left as future work.

%% file: sections/acknowledgments.tex
This research was supported by the King's Computational Research,
Engineering and Technology Environment (CREATE) at King's College
London (\url{https://doi.org/10.18742/rnvf-m076}), which provided the
computational resources used for the experiments reported in this paper.

%% file: appendix/appendix.tex
\section{Additional experiments}
\label{app:additional}

\subsection{Additional BP-free Baselines on ImageNet-100}
\label{app:additional_baselines}

We additionally evaluated DeeperForward~\citep{deeperforward} on ImageNet-100 by adapting its published per-block goodness configuration and training protocol to our VGG-16 architecture and data pipeline. DeeperForward achieves $13\%$ Fusion accuracy on IN-100, compared with $59.38\%$ for our ASGE reproduction and $73.01\%$ for our full BiCovG + FAL model. The large gap reflects DeeperForward's architectural constraints rather than a fundamental limitation of BP-free learning; we include this result for completeness and do not include it in the main comparison table due to the absence of a matched Tiny-ImageNet evaluation.

\subsection{Full Architecture Configuration}
\label{app:arch}

Table~\ref{tab:arch_full} provides the complete layer-wise configuration of the 16-block VGG-16-derived network used in all experiments. All convolutions use $3{\times}3$ kernels with padding 1 and stride 1. RMSPool (kernel $2{\times}2$, stride 2) halves spatial dimensions at layers 1, 3, 7, and 11. ImageNet-100 ($224^2$) inputs are first downsampled via $\text{AvgPool}(2{\times}2)$ to $112{\times}112$ before layer 0.

\begin{table}[h]
\centering
\small
\caption{Layer-wise architecture configuration. RMSPool ($\checkmark$) halves spatial dimensions.}
\label{tab:arch_full}
\begin{tabular}{ccccc}
\toprule
Layer $l$ & Channels (in$\to$out) & RMSPool & Spatial (Tiny-IN $64^2$) & Spatial (IN-100 $224^2$) \\
\midrule
0  & $3\to128$     &              & $64$          & $112$        \\
1  & $128\to128$   & $\checkmark$ & $64\to32$     & $112\to56$   \\
2  & $128\to256$   &              & $32$          & $56$         \\
3  & $256\to256$   & $\checkmark$ & $32\to16$     & $56\to28$    \\
4  & $256\to512$   &              & $16$          & $28$         \\
5  & $512\to512$   &              & $16$          & $28$         \\
6  & $512\to512$   &              & $16$          & $28$         \\
7  & $512\to512$   & $\checkmark$ & $16\to8$      & $28\to14$    \\
8  & $512\to512$   &              & $8$           & $14$         \\
9  & $512\to512$   &              & $8$           & $14$         \\
10 & $512\to512$   &              & $8$           & $14$         \\
11 & $512\to512$   & $\checkmark$ & $8\to4$       & $14\to7$     \\
12 & $512\to512$   &              & $4$           & $7$          \\
13 & $512\to512$   &              & $4$           & $7$          \\
14 & $512\to512$   &              & $4$           & $7$          \\
15 & $512\to512$   &              & $4$           & $7$          \\
\bottomrule
\end{tabular}
\end{table}

For CIFAR-100 ($32{\times}32$), we use a shallower 8-block VGG-8 backbone with block configuration $(4,4)$, i.e., the FAL is inserted between block 3 and block 4. All other per-block components --- Conv$_{3\times3}\to$BN$\to$ReLU, optional RMSPool, BiCovG on post-ReLU pre-pool features, $W_l$ readout, dropout $p{=}0.1$ at block end, and inter-block detach (or inter-group detach in the FAL setting) --- match the VGG-16 configuration.

\begin{table}[h]
\centering
\small
\caption{VGG-8 layer-wise configuration for CIFAR-100. The FAL is inserted between block 3 and block 4 (group split $4|4$).}
\label{tab:arch_vgg8}
\begin{tabular}{cccc}
\toprule
Block $l$ & Channels (in$\to$out) & RMSPool & Spatial (CIFAR-100 $32^2$) \\
\midrule
0 & $3\to128$    &              & $32$        \\
1 & $128\to256$  & $\checkmark$ & $32\to16$   \\
2 & $256\to256$  &              & $16$        \\
3 & $256\to512$  & $\checkmark$ & $16\to8$    \\
4 & $512\to512$  & $\checkmark$ & $8\to4$     \\
5 & $512\to512$  & $\checkmark$ & $4\to2$     \\
6 & $512\to512$  &              & $2$         \\
7 & $512\to512$  &              & $2$         \\
\bottomrule
\end{tabular}
\end{table}

\subsection{Full Training Configurations}
\label{app:training}

\begin{table}[h]
\centering
\small
\caption{Training hyperparameters for Tiny-ImageNet and ImageNet-100.}
\label{tab:train_tin_in100}
\begin{tabular}{ll}
\toprule
Hyperparameter & Value \\
\midrule
Optimizer      & SGD (momentum $= 0.9$, weight decay $= 10^{-4}$) \\
LR schedule    & Cosine annealing $0.05 \to 5{\times}10^{-4}$ \\
Batch size (Tiny-IN) & 128/GPU, DDP $2\times$GPU (eff.\ 256) \\
Batch size (IN-100)  & 64/GPU, DDP $2\times$GPU (eff.\ 128) \\
Epochs         & 90 \\
Augmentation   & RandomCrop + HFlip + ColorJitter + ImageNet normalize \\
               & (IN-100: RandomResizedCrop(224)) \\
AMP            & Enabled \\
Gradient clip  & $\|\nabla\|_2 \leq 1.0$ \\
Dropout        & 0.1 \\
HGB warmup\textsuperscript{$\dagger$} & 5 epochs (linear) \\
\bottomrule
\end{tabular}
\vspace{2pt}
{\small $^\dagger$HGB configurations only; BP-free configurations use no warmup.}
\end{table}

\begin{table}[h]
\centering
\small
\caption{Training hyperparameters for CIFAR-100 (VGG-8, 8-block).}
\label{tab:train_cifar}
\begin{tabular}{ll}
\toprule
Hyperparameter & Value \\
\midrule
Optimizer      & AdamW (weight decay $= 10^{-3}$) \\
LR schedule    & Cosine annealing $2{\times}10^{-4} \to 10^{-5}$ \\
Batch size     & 128 \\
Epochs         & 400 \\
Augmentation   & RandomCrop(32, pad=4) + HFlip + ColorJitter(0.4) + RandomGrayscale(0.2) \\
AMP            & Disabled \\
Gradient clip  & — \\
\bottomrule
\end{tabular}
\end{table}

\subsection{Interleaved Training for Memory Efficiency}
\label{app:interleaved}

Since blocks are decoupled by \texttt{detach()}, their forward-backward passes are mathematically independent. This enables an \textbf{interleaved execution} strategy that reduces peak activation memory from $O(L)$ to $O(1)$:
\begin{equation}
    \text{For } l = 0, \ldots, L{-}1: \quad
    \text{forward}(\theta_l) \;\to\; \mathcal{L}_l \;\to\; \text{backward}(\theta_l) \;\to\; \text{step}(\theta_l) \;\to\; \text{release activations}
\end{equation}
At any given time, only one block's activation graph resides in GPU memory. This is mathematically equivalent to the standard greedy forward pass (Section~\ref{sec:method:training}) but trades sequential computation for a constant activation memory footprint (parameter memory is unchanged). This optimization is orthogonal to GPU count and block configuration — it applies to any variant (BP-free, HGB) on any hardware setup.